\documentclass{article}

\usepackage{microtype}
\usepackage{graphicx}
\usepackage{subfigure}
\usepackage{booktabs} 
\usepackage{natbib}
\usepackage{amsmath}
\usepackage{amsfonts}
\usepackage{amssymb}
\usepackage{xcolor}

\usepackage{hyperref}

\usepackage[accepted]{mlsys2022}

\begin{document}

\twocolumn[
\mlsystitle{Transformer Acceleration with Dynamic Sparse Attention}



\mlsyssetsymbol{equal}{*}

\begin{mlsysauthorlist}
\mlsysauthor{Liu Liu}{equal,cs,ece}
\mlsysauthor{Zheng Qu}{equal,ece}
\mlsysauthor{Zhaodong Chen}{ece}
\mlsysauthor{Yufei Ding}{cs}
\mlsysauthor{Yuan Xie}{ece}
\end{mlsysauthorlist}

\mlsysaffiliation{cs}{Department of Computer Science, University of California, Santa Barbara, USA}
\mlsysaffiliation{ece}{Department of Electrical and Computer Engineering, University of California, Santa Barbara, USA}
\mlsyscorrespondingauthor{Liu Liu}{liu\_liu@ucsb.edu}

\mlsyskeywords{Machine Learning, Efficient Methods}

\vskip 0.3in

\begin{abstract}
Transformers are the mainstream of NLP applications and are becoming increasingly popular in other domains such as Computer Vision. Despite the improvements in model quality, the enormous computation costs make Transformers difficult at deployment, especially when the sequence length is large in emerging applications. Processing attention mechanism as the essential component of Transformer is the bottleneck of execution due to the quadratic complexity. Prior art explores sparse patterns in attention to support long sequence modeling, but those pieces of work are on static or fixed patterns. We demonstrate that the sparse patterns are dynamic, depending on input sequences. Thus, we propose the Dynamic Sparse Attention (DSA) that can efficiently exploit the dynamic sparsity in the attention of Transformers. Compared with other methods, our approach can achieve better trade-offs between accuracy and model complexity. Moving forward, we identify challenges and provide solutions to implement DSA on existing hardware (GPUs) and specialized hardware in order to achieve practical speedup and efficiency improvements for Transformer execution.
\end{abstract}
]



\printAffiliationsAndNotice{\mlsysEqualContribution} 

\section{Introduction}
Transformers \citep{Vaswani2017attention} have become the driving force for sequence modeling tasks such as neural machine translation \citep{ott2018scaling}, language understanding \citep{devlin2019bert}, and generative modeling \citep{parmar2018image, brown2020language}. Equipped with the self-attention mechanism, Transformers are capable of handling long-range dependencies. 

Despite the impressive progress made by Transformers, the computational requirements make the deployment of Transformer-based models difficult at inference time, especially when processing long sequences. The quadratically scaled self-attention modules are the execution bottleneck under long sequences. Therefore, many studies propose Transformer variants to mitigate the quadratic time and space complexity issue. Some approaches are primary for memory footprint reduction during training while efficient inference is being understudied \citep{Gong2019Stacking,dai2019transformerxl,Kitaev2020Reformer,roy2021efficient}. Other methods use fixed or static sparse attention patterns to save computations \citep{child2019generating,qiu2020blockwise,beltagy2020longformer,zaheer2020big}. However, we find that intrinsic sparse patterns in attention are naturally dynamic, depending on input sequences. Thus, we propose to exploit the dynamic sparse patterns to save attention computations without sacrificing the representation power of attention. Intuitively, posing static sparsity constraints in attention could be too strong to capture dynamic attention connections.


We propose the Dynamic Sparse Attention (DSA) approach that exploits dynamic sparsity to improve efficiency. The challenge is to efficiently search for sparse patterns close to \textit{oracle} sparse patterns that keep all the important attention weights.
We formulate the searching as a prediction problem and augment the standard attention mechanism with a prediction path. As discussed in Section \ref{sec:method}, we first obtain an approximation of attention scores with low computational costs. Then, we predict the sparse attention patterns using the approximate attention scores. With the predicted sparse attention patterns represented as binary masks, we can save computations involved in full attention scores, $softmax$, and attention outputs.

Compared with static sparse attention methods, our method is dynamic and naturally captures sparse attention patterns of different input sequences. We observe important tokens that attract a large portion of attention weights from other tokens, similar to the global attention method \cite{beltagy2020longformer,zaheer2020big}. However, the positions of global tokens are input-dependent, and our method can effectively identify such varieties, instead of relying on domain knowledge to predetermine certain global tokens in fixed positions. Compared with other low-rank approximation methods, the approximation in DSA is only for sparsity prediction without strict and static constraints on attention positions. Therefore, our method can maintain the representation power of full attention while reducing unnecessary attention weights.

Although DSA can save theoretical computations and maintain attention capability, achieving practical speedups and energy savings on real hardware systems is challenging. We discuss the implications of DSA on existing GPU architectures and specialized hardware accelerators. We extend the fine-grained dynamic sparsity as searched by DSA to structural dynamic patterns, such as block-wise and vector-wise. We give the study on structural sparse patterns vs. attention's expressive power and explore the opportunities for dataflow optimization and data reuse from dynamic sparsity. 

Our evaluation in Section \ref{sec:eval} shows that DSA can achieve 95\% sparsity in attention weights without compromising model accuracy. Under this setting, the overall computational saving is up to 4.35$\times$ compared with full attention, while the sparsity prediction only introduces around 1.17\% to 1.33\% computational overhead. Experimental results on NVIDIA V100 GPU show that, under 90\% sparsity ratio, applying vector-wise sparsity on DSA delivers $1.15\times$ speedup on attention score computation, $14.6\times$ speedup on softmax computation, and $1.94\times$ speedup on attention output computation, with only 0.1\% of accuracy loss. Finally, through hardware specialization, we can further explore token-level parallelism and computation reordering for DSA. Our characterization results show that this can reduce the total memory access of attention computation by up to $2.54\times$.

\section{Background and Motivation} \label{sec:motiv}
Before we describe our method in detail, we first introduce the preliminaries of the standard attention mechanism used in vanilla Transformers. Then, we discuss the challenge of serving long sequences under the quadratic complexity of attention. Finally, we demonstrate that redundancy exists in attentions and dynamic sparse patterns are naturally expressed in attention.

\subsection{Preliminaries of Attention}
The attention mechanism is the essential component of Transformers \citep{Vaswani2017attention}. Self-attention operates on input representations of length $l$, $X \in \mathbb{R}^{l \times d}$, with three linear projections namely, query, key, and value as
\begin{equation}
    Q, K, V = X W_Q, X W_K, X W_V
\end{equation}
, where $Q \in \mathbb{R}^{l \times d_k}$ denotes the queries, $K \in \mathbb{R}^{l \times d_k}$ denotes the keys, and $V \in \mathbb{R}^{l \times d_v}$ denotes the values. After linear projections, the attention weights $A \in \mathbb{R}^{l \times l}$ is defined as
\begin{equation} \label{eq:attention}
    A = \phi(\frac{QK^\top}{\sqrt{d_k}})
\end{equation}
where $\phi$ is the row-wise $softmax(\cdot$) function. Finally, the output values are computed by multiplying the attention weights $A$ with the projected values $V$ as
\begin{equation} \label{eq:output}
    Z = AV.
\end{equation}

Serving Transformer-based models is challenging when the input sequence length $l$ is large. 
When using long sequences, computing Eq. (\ref{eq:attention}) and Eq. (\ref{eq:output}) consumes the majority of operations and becomes the bottleneck of model evaluation. The asymptotic complexity of attention $O(l^2 d_k + l^2 d_v)$ is quadratic to sequence length $l$.

\subsection{Intrinsic Sparsity in Attention Weights}
A number of efficient Transformer variants have been proposed to mitigate the quadratic complexity of self-attention \cite{child2019generating,beltagy2020longformer,zaheer2020big,shi2021sparsebert}. One straightforward way to exploit the intrinsic redundancy in attention is forming sparse patterns as in
\begin{equation} \label{eq:mask}
    A = \phi(QK^\top - c(1 - M)),
\end{equation}
where $M \in \{0, 1\}^{l \times l}$ represents the sparse attention pattern, $c$ is a large constant ($1e^4$) such that where $M_{ij} = 0$, indicating unimportant attention, $A_{ij} = 0$ after $softmax$ normalization. Here, we omit $\sqrt{d_k}$ for simplicity. The sparse patterns can be pre-determined into global, block, random, or a combination of different patterns. Another way to determine sparse patterns is through trainable masks. However, all these methods explore static or fixed sparse patterns, restricting viable attention connections.

\subsection{Dynamic Sparse Patterns in Attention}
A common motivation of sparse attention methods is that not all attention weights, i.e., probabilities, are equally important in Eq. (\ref{eq:output}). A large portion of attention weights do not contribute to attention output and are redundant. In other words, only a small portion of attention weights are useful. However, we find that sparse patterns in attention are inherently dynamic and data-dependent.


\begin{figure}[t]
    \centering
    \vspace{-8pt}
    \includegraphics[width=0.47\textwidth]{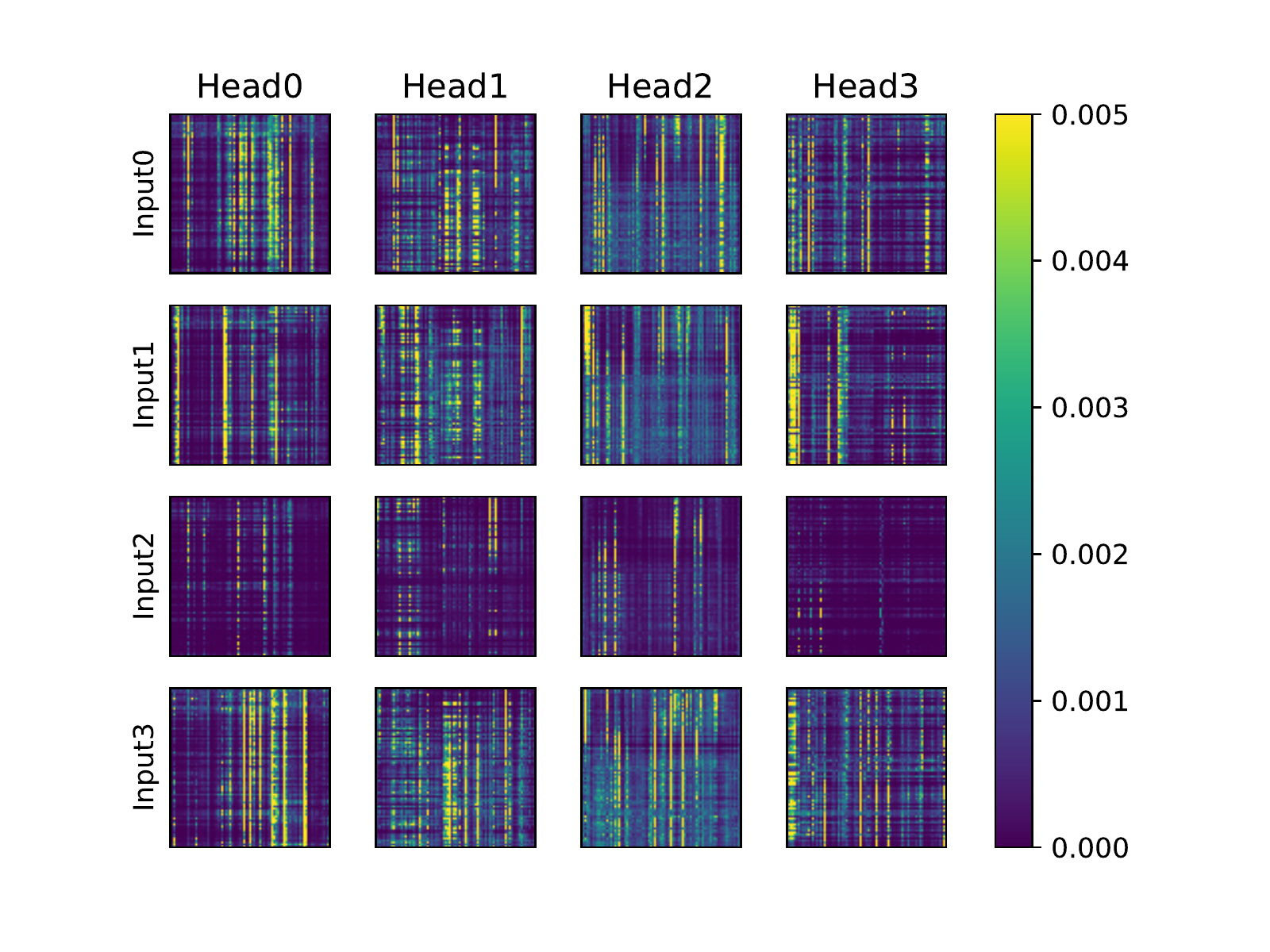}
    \vspace{-25pt}
    \caption{Visualization of attention weights from different inputs and attention heads. Only a small amount of attention weights are important. Note values $> 0.005$ are clamped to show as $0.005$.}
    \label{fig:probs}
    \vspace{-15pt}
\end{figure}

Here, we further support our hypothesis by showing the original attention weights matrix (after $softmax$ normalization) in Figure~\ref{fig:probs}. The model used here is a vanilla Transformer and the benchmark is Text Classification from Google Long-Range Arena\citep{long-range-arena}. 
Figure~\ref{fig:probs} indicates that only a small amount of attention weights are with large magnitude and a significant portion is near zero. We want to emphasize that this shows the raw attention weights without forcing any sparsity constraints or fine-tuning, which indicates that redundancy naturally exists in attention. In short, attention mechanism exhibits the focused positions on a set of important tokens. 

More importantly, the attention weights have dynamic sparse patterns. As shown in Figure \ref{fig:probs}, the sparse patterns in attention weights are dynamically changing depending on the input sequence. Different heads in multi-head attention also have different sparse patterns. The characteristic of dynamic sparsity in attention weights motivates us to explore effective methods to eliminate the redundancy and save computations. Prior work on static or fixed sparse patterns cannot capture the dynamically changing attention weights.
\section{Dynamic Sparse Attention} \label{sec:method}
From Section \ref{sec:motiv}, we show that attention weights have intrinsic sparse patterns, and the positions of important attention weights are dynamically changing as different input sequences. While attention exhibits dynamic sparse patterns, how to efficiently and effectively obtain the dynamic sparse patterns remains challenging. We formulate the process of identifying sparse attention patterns as a prediction problem. The key challenge is how to obtain an approximate attention predictor that can accurately find the sparse patterns while keeping the prediction overhead small.

\begin{figure}[t]
    \centering
    \includegraphics[width=0.45\textwidth]{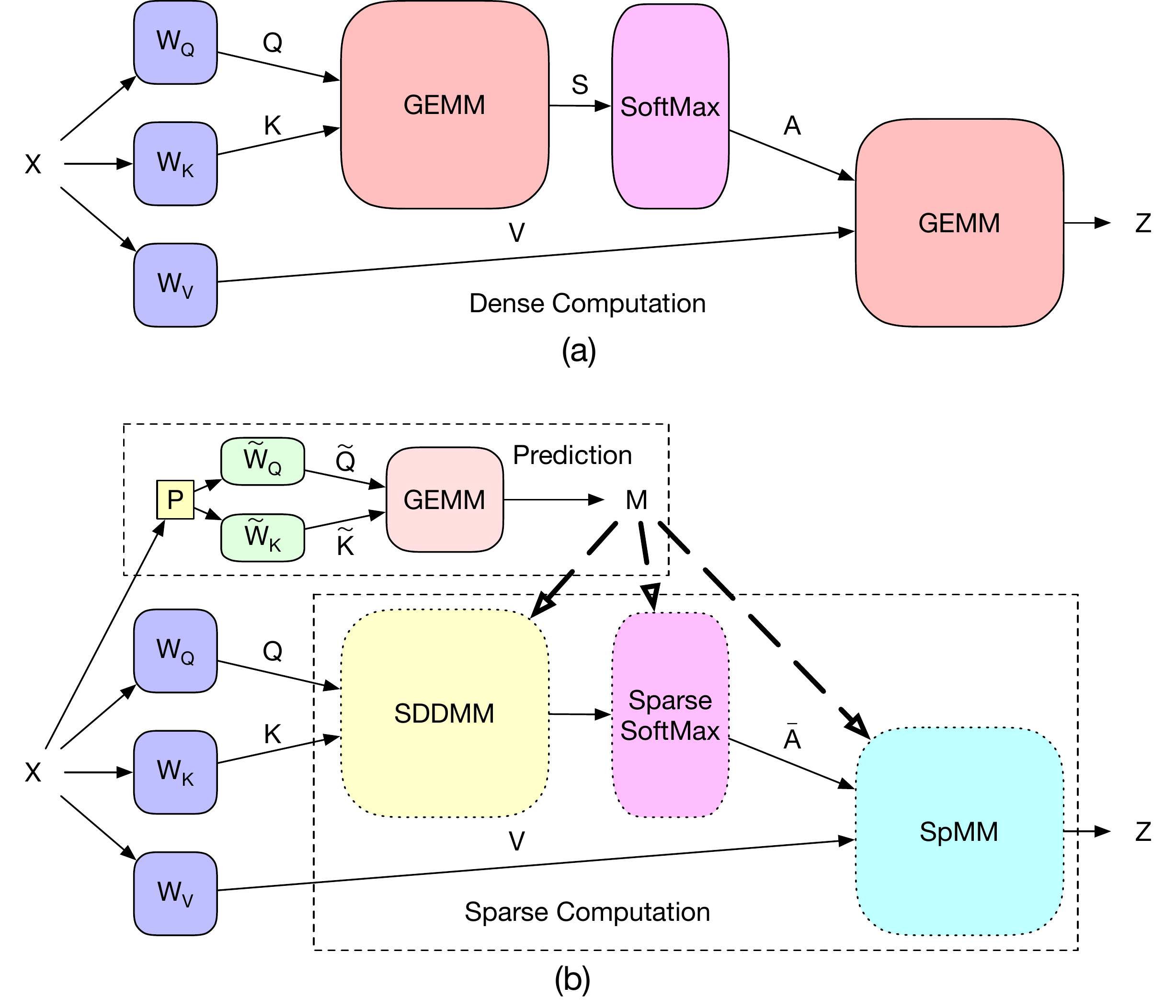}
    \vspace{-15pt}
    \caption{(a) Standard full attention; (b) Dynamic sparse attention with approximation-based prediction and sparse computation.}
    \label{fig:method}
    \vspace{-10pt}
\end{figure}

Here, we present \textit{Dynamic Sparse Attention} (DSA) that exploits sparsity in attention weights to reduce computations.
The principle of our method is to effectively search for dynamic sparse patterns without enforcing strict and static constraints on attention while keeping the searching cost small. Our approach leverages trainable approximation to predict sparse attention patterns. As shown in Figure \ref{fig:method}, we use a prediction path based on low-rank transformation and low-precision computation. The prediction path processes input sequences functionally similar to query and key transformations but at much lower computational costs. Given the prediction results that approximate $Q K^\top$ well, we can search sparse patterns based on the magnitude of prediction results.

\subsection{Design of Prediction Path}
We denote attention scores as $S = Q K^\top$ and omit the scaling factor for simplicity. As shown in Figure \ref{fig:method}(a), two general matrix-matrix multiplication kernels (GEMM) and one $softmax$ kernel consume the majority of computations in self-attention. 
We construct a pair of approximate query and key transformations in the prediction path to compute for approximate score $\tilde{S}$, as in
\begin{equation}
    \tilde{Q}, \tilde{K} = XP \tilde{W_Q}, XP \tilde{W_K}.
\end{equation}
Here $P \in \sqrt{\frac{3}{k}}\cdot \{-1, 0, 1\}^{d \times k}$ is a sparse random projection matrix shared by both paths, and $\tilde{W_Q} \in \mathbb{R}^{k \times k}, \tilde{W_K} \in \mathbb{R}^{k \times k}$ are parameters in approximating query and key.

Then, we have approximate attention scores $\tilde{S} = \tilde{Q}\tilde{K}^\top$. From $\tilde{S}$, we can predict sparse attention masks $M$ using thresholds, where the threshold values are either fixed by tuning from the validation set or determined by $top-k$ searching. When $\tilde{S}$ is well approximated with accurate attention scores $S$, the large scores in $\tilde{S}$ are also large in $S$ with high probability. The resulting sparse attention weights $\bar{A}$ is used to multiply the value matrix $V$ similar to Eq. \ref{eq:output}.


\textbf{Optimization of Approximation.} The random projection matrix $P$ is constant after initialization and shared by two approximate transformations. We obtain the trainable parameters, $\tilde{W_Q}$ and $\tilde{W_K}$, through minimizing the mean squared error (MSE) as the criterion to optimize for approximation:
\begin{equation}
    L_{MSE} = \frac{1}{B} || S - \tilde{S} ||_2^2 = \frac{1}{B} || QK^\top - \tilde{Q}\tilde{K}^\top ||_2^2
\label{equ:loss_fn}
\end{equation}
where B is the mini-batch size. 

Given the motivation of finding dynamic sparse patterns, the hypothesis of our method is that there exist \textit{oracle} sparse patterns that perform well. Such that the optimization target is to approximate full attention scores $S$ well enough to predict sparse patterns. We further give the results of applying oracle sparse patterns by directly dropping small-magnitude attention weights during inference without fine-tuning the model. As listed in Table \ref{tab:qa_sparse}, around 90\% (up to 97\%) of small attention weights can be dropped with negligible accuracy loss.




\begin{table}[!ht]
\vspace{-15pt}
    \centering
    \caption{Sparsity in attention weights, where values $< \theta$ are set to zero. A significant portion of attention weights that have small magnitude are redundant. The accuracy metrics are Exact Match (EM) and F1 Score.}
    \begin{tabular}{c|c|c|c} \hline \hline
        Case & Sparsity & EM & F1 \\ \hline
        Base & 0\% & 81.49 & 88.70 \\
        $\theta = 0.001$ & 75\% - 95\% & 81.50 & 88.70 \\
        $\theta = 0.01$ & 94\% - 97\% & 80.51 & 87.85 \\ \hline
    \end{tabular}
    \vspace{-15pt}
    \label{tab:qa_sparse}
\end{table}

\subsection{Model Adaptation} 
When sparse attention scores are masked out to generate sparsity in attention, the remaining attention weights, i.e., the important weights, are scaled up as the denominator becomes small. Leaving the disturbed attention weights intact will degrade model quality. As a countermeasure, we propose to fine-tune model parameters with dynamic sparse constraints, referred to as model adaptation. With adaptation, the model evaluation accuracy can recover to be on par with full attention baselines, while the computational costs are significantly reduced. 

We do not change the computational graph and the loss function of the original model, except adding dynamic sparse constraints in attention as mask $M$. As a result, the new attention $\bar{A}$ are sparse and only have important weights from prediction. Given a pre-trained model, our method jointly fine-tunes the model parameters and parameters of the prediction path as in
\begin{equation}
    L = L_{Model} + \lambda L_{MSE}
\end{equation}
where $\lambda$ is the regularization factor of MSE. Our method can also train from scratch with initialized model parameters.


Our method approximates the original attention score with a low-rank matrix $\tilde{S}$. When training the model with loss function in Eq. \ref{equ:loss_fn}, the gradient from $L_{MSE}$ will be passed to both the low-rank approximation $\tilde{S}$ and the original attention score $S$. Intuitively, this loss function not only makes $\tilde{S}$ a better approximation of $S$, but also makes $S$ easier to be approximated by a low-rank matrix, i.e., by reducing the rank of $S$. On the other hand, the loss $L_{Model}$ guarantees the rank of $S$ to be high enough to preserve the model accuracy. In other words, the joint optimization of $L_{Model}$ and $L_{MSE}$ implicitly learns a low-rank $S$ with a learnable rank depending on the difficulty of the task. Our design brings two advantages. First, the rank of $S$ will be automatically adjusted to tasks with different difficulty levels. Hence, our method can potentially achieve higher accuracy on difficult tasks and higher speedup on simple tasks compared with low-rank approximation methods using fixed rank. Second, as the rank of $\tilde{S}$ only implicitly influences the rank of $S$, the final result is less sensitive to the hyper-parameter $k$. 


\subsection{Computation Saving Analysis}
DSA introduces additional computations in the prediction step, but the overall computation saving from sparse attention kernels is fruitful and can have practical speedup. The original full attention takes $O(l^2 d_k + l^2 d_v)$ MACs (multiply-and-accumulate operations) asymptotically. However, the asymptotic analysis does not consider practical concerns such as sparsity, quantization, and data reuse. Here, we augment the traditional asymptotic analysis with a sparsity factor $\alpha$ and a quantization factor $\beta$. In this way, DST prediction takes $O(\beta l d_k k + \beta l^2 k )$ MACs; DST attention takes $O(\alpha l^2 d_k + \alpha l^2 d_v)$ MACs. Both $\alpha$ and $\beta$ are determined depending on tasks and underlying hardware platforms. In our settings, we choose $\alpha$ between 90\% and 98\% and our GPU kernels can achieve practical speedups. We assume the baseline model uses FP32 as the compute precision and set prediction precision to be INT4. 
The execution time on $softmax$ is not revealed in asymptotic analysis but is one of the major time-consuming components. Our method can also save the time of $softmax$ kernel with the same sparse attention patterns.

\subsection{Implications for Efficient Deployment}
Compared with standard attention, DSA exhibits two new features that can potentially affect model deployment. Firstly, a light-weight prediction path is attached to the attention layer to search for dynamic sparse patterns. The prediction involves approximation of attention scores, which is essentially a low-precision matrix-matrix multiplication (GEMM). While NVIDIA GPUs with Tensor Cores support data precision as low as INT8 and INT4, DSA prediction can tolerate INT2 computation on certain benchmarks. Therefore, specialized hardware is preferable when seeking ultra-efficient attention estimation. In Section~\ref{sec:hardware}, we introduce two types of architectures to support multi-precision computations. 

Secondly, the predicted sparse patterns can be used to reduce unnecessary attention computations. In other words, instead of computing $QK^\top$ and $AV$ as two dense GEMM operations, we can reformulate $QK^\top$ as a sampled dense dense matrix multiplication (SDDMM) and $AV$ as a sparse matrix-matrix multiplication (SpMM). When processing SDDMM and SpMM kernels on GPU, data reuse is the key disadvantage that limits its performance compared with GEMM. Therefore, we extend DSA to support structural sparsity that can improve the data reuse of both SDDMM and SpMM kernels. We implement customized kernels that take advantage of the sparsity locality to improve kernel performance, achieving practical runtime speedup on NVIDIA V100 GPU. Also, we demonstrate our choice of structural sparsity pattern and that DSA is able to maintain the model expressive power with the extra constraints. 

As for specialized hardware, the advantage of DSA can be fully exploited as the specialized architecture and dataflow is able to deal with fine-grained sparsity, therefore achieving optimal sparsity ratio and computation reduction. However, the challenge also arises as irregular sparsity causes load imbalance and under-utilization of processing elements. Moreover, instead of independently executing SDDMM and then SpMM, we point out that more optimization opportunities can be explored when considering the whole process as a two-step SDDMM-SpMM chain. Please refer to Section~\ref{sec:hardware} for more architectural design details and experimental results.

\section{Evaluation} \label{sec:eval}
In this section, we evaluate the performance of DSA over representative benchmarks from Long-Range Arena~\citep{long-range-arena}. We first compare the model accuracy of DSA with dense vanilla transformers and other efficient transformer models. Then, we present a sensitivity study over different configurations of the prediction path. By choosing different number of prediction parameters, DSA is able to achieve flexible trade-offs between computational cost and model accuracy. Finally, we study the model efficiency of DSA by analyzing the computational cost (MACs) and relative energy consumption. 

\subsection{Experiment Settings}\label{sec:exp_setting}
The datasets used are from Long-Range Arena (LRA), which is a benchmark suite for evaluating model quality under long-sequence scenarios. In LRA, different transformer models are implemented using Jax~\citep{jax2018github} API and optimized with just-in-time ($jax.jit$) compilation. We implement DSA on top of the vanilla transformer provided by LRA and compare it with other models included in LRA. Specifically, the self-attention layer in the vanilla transformer is augmented by the DSA method as described in Section \ref{sec:method}. All the other model configurations are kept the same for a fair comparison.

We incorporate three tasks from the LRA benchmark in our experiment, including Text Classification, Document Retrieval, and Image Classification. The Long ListOps and Pathfinder tasks are excluded. We provide benchmark descriptions and experiment configurations in Appendix \ref{app:descript}.

\subsection{Accuracy} 
\begin{figure}[t]
  \centering
   \includegraphics[width=0.48\textwidth]{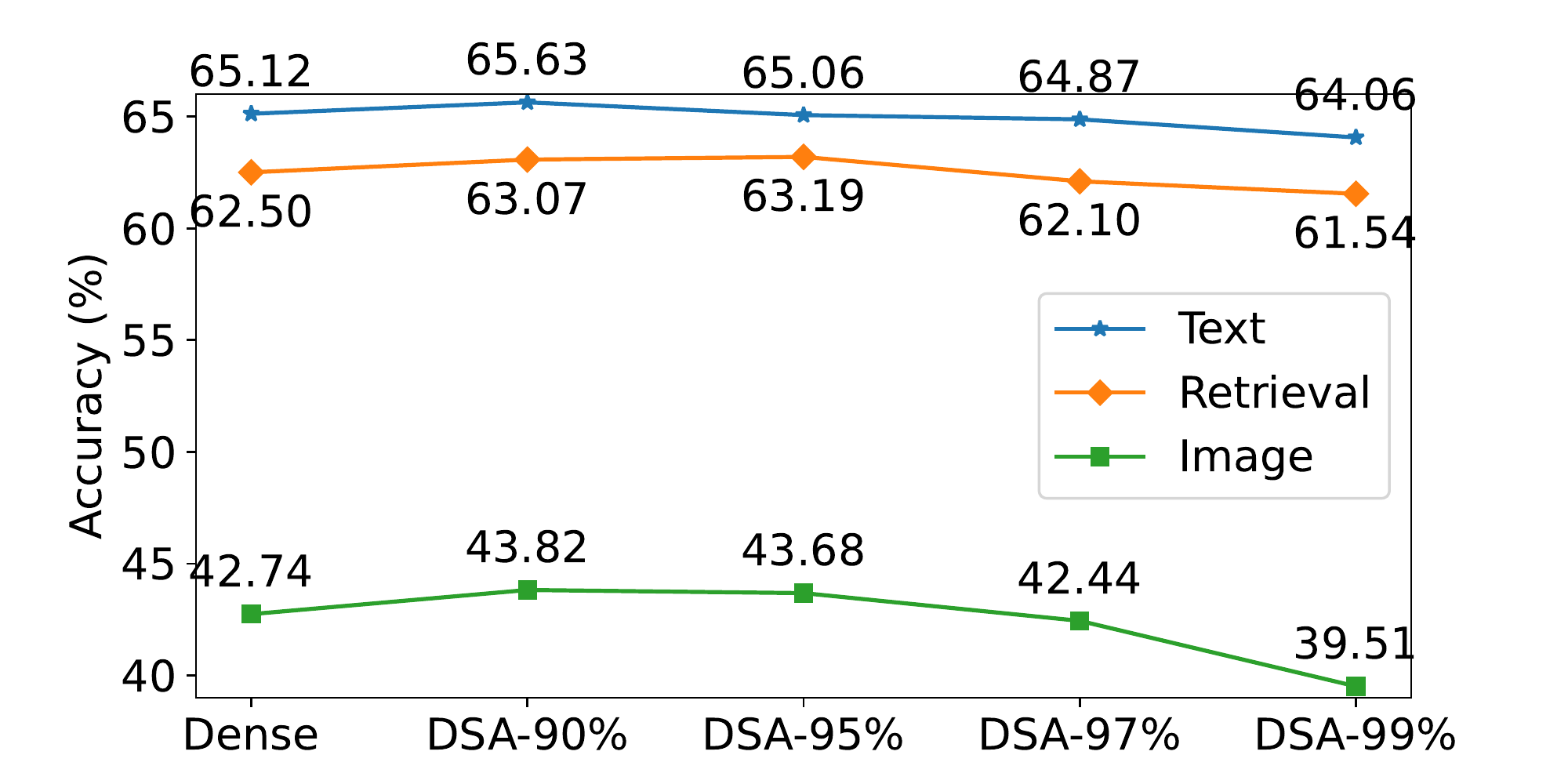}
   \vspace{-20pt}
  \caption{Overall model accuracy of DSA (\textbf{fine-tuned} from a pretrained checkpoint) compared with vanilla dense transformer.}
  \label{Overall-Accu}
  \vspace{-10pt}
\end{figure}

Figure~\ref{Overall-Accu} presents the overall model accuracy of DSA on different LRA tasks. In this experiment, the DSA model is fine-tuned from a pretrained vanilla transformer by jointly updating the model parameters and prediction parameters using the combined loss of $L_{MSE}$ and $L_{Model}$.
Different percentage numbers indicate the sparsity ratio that we applied to the DSA models. For instance, \textbf{DSA-90\%} means that we only keep 10\% of the attention weights in each row of the attention matrix, while masking out all the other 90\% of the weights. The sparsity ratio constraint is uniform for all the heads and attention layers in the DSA model.

As shown in Figure~\ref{Overall-Accu}, for all the evaluated tasks, dense transformer possesses a considerable amount of redundancy in the attention matrix under the long-sequence condition, which supports our previous claim in Section~\ref{sec:motiv}. Specifically, we can safely mask out up to 95\% of the attention weights without suffering from any accuracy degradation. In fact, by jointly optimizing the model parameters to adapt dynamic sparse attention, DSA delivers slightly higher performance with 90\% and 95\% sparsity ratio. Even with up to 99\% of sparsity, DSA still demonstrates promising performance with negligible accuracy drop compared with the dense baseline.

\begin{table}
\vspace{-3pt}
  \caption{Accuracy of different Transformer models on the LRA benchmark suite \cite{long-range-arena}. For a fair comparison, we follow the instructions in LRA and train our model \textbf{from scratch}. \textbf{DSA-90\%} uses projection scale $\sigma=0.25$ and INT4 quantization.}
  \vspace{2pt}
  \label{accuracy-comp}
  \centering
  \begin{tabular}{c|c|c|c|c}
    \toprule
   Model & Text  & Retrieval  & Image  & Avg\\
    \midrule
    Transformer & 65.12 & \underline{62.5} & 42.74  & 56.79    \\
      \midrule
    Local Attention   & 52.98 & 53.39 &  41.46 & 50.89      \\
    Sparse Trans.     &  63.58  & 59.59 & \textbf{44.24} & \underline{55.80}  \\
    Longformer    	&62.85 & 56.89 & 42.22 & 53.99 \\
    Linformer     		&  53.94 &  52.27 & 38.56 & 48.26  \\
   Reformer     		& 56.10& 53.40 & 38.07 & 49.19  \\
    Sinkhorn Trans.   &61.20& 53.83 & 41.23 & 52.09 \\
    Synthesizer     	& 61.68 & 54.67 &  41.61 & 52.65  \\
    BigBird  			 & 64.02 & 59.29 & 40.83 & 54.71 \\
    Linear Trans.    & \textbf{65.90} & 53.09 & 42.34 & 53.78  \\
    Performer 			& 65.40 & 53.82 & 42.77  & 54.00    \\
   \midrule
      \textbf{DSA-90\%} 		& \underline{65.62} & \textbf{63.07} & \underline{43.75}  & \textbf{57.48}    \\
    \bottomrule
  \end{tabular}
  \vspace{-5pt}
\end{table}

To fairly compare with other transformer variants provided by LRA, we follow several training constraints during our experiment. For example, instead of fine-tuning from a pretrained baseline, the DSA model used in the comparison is obtained from a randomly initialized model, i.e., training from scratch. We also fix other model parameters (e.g., number of layers, number of heads, hidden dimension) and training configurations (e.g., total training steps). The results are shown in Table~\ref{accuracy-comp}. We use \textbf{DSA-90\%} with quantization precision to be INT4, and let the random projection dimension scale $\sigma$=$k/d$=$0.25$. As we can see from the table, DSA achieves first-tier performance in all three tasks and delivers a leading average score on the LRA benchmarks.

\begin{figure}[!ht]
\vspace{-8pt}
  \centering
  \includegraphics[width=0.48\textwidth]{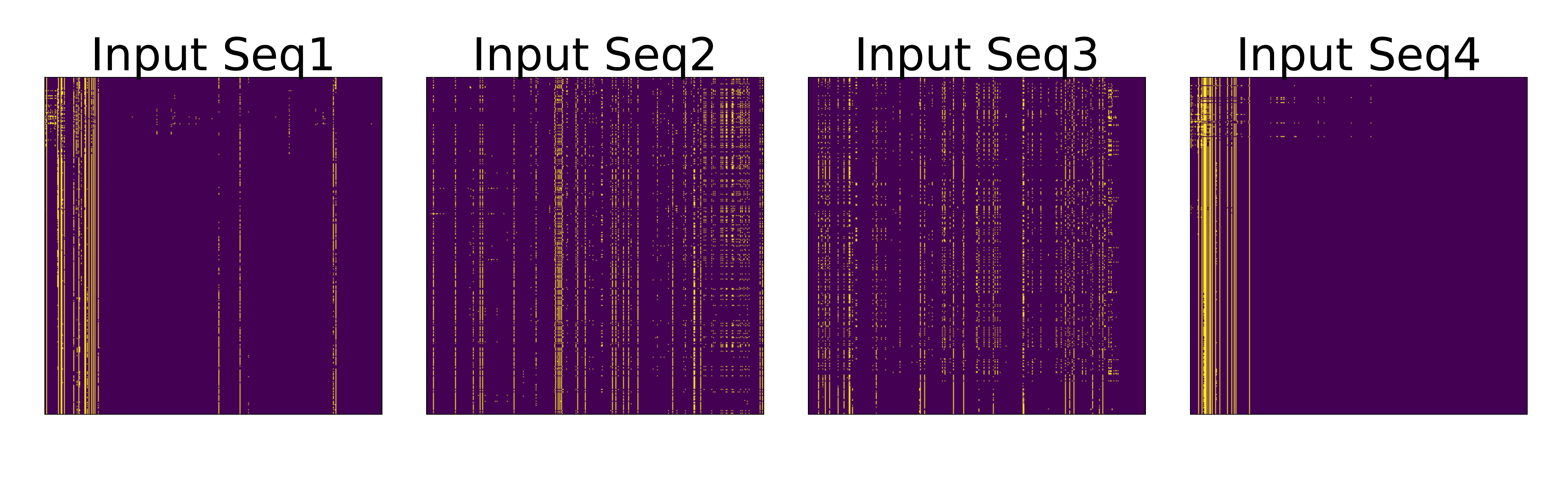}
  \vspace{-30pt}
  \caption{\textit{Oracle} attention mask generated by \textit{top-k} selection.}
  \label{fig:oracle}
  \vspace{-10pt}
\end{figure}

\begin{figure}[!ht]
\vspace{-5pt}
  \centering
  \includegraphics[width=0.48\textwidth]{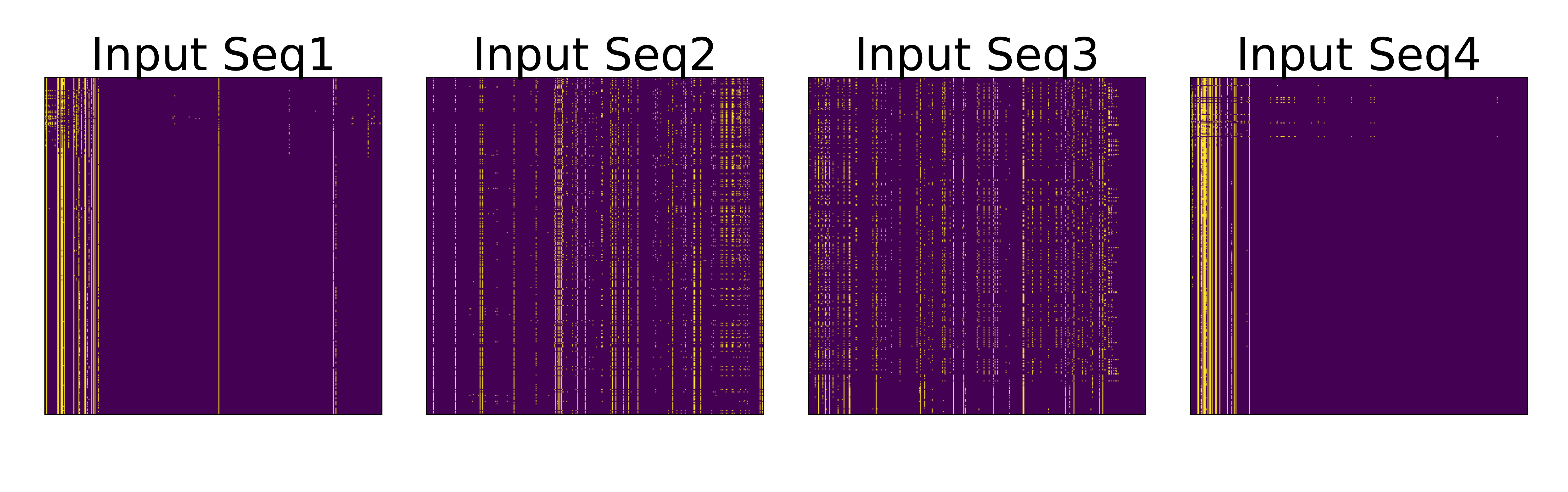}
    \vspace{-30pt}
  \caption{Sparse attention mask generated by DSA prediction.}
  \label{fig:dsa_pattern}
  \vspace{-5pt}
\end{figure}

This encouraging performance mainly comes from two aspects. Firstly, joint optimization ensures that the DSA model can well adapt to the sparse attention patterns for computing the attention output. Secondly, the trainable prediction path is able to accurately capture the input-dependent patterns. 
Figure \ref{fig:oracle} shows the \textit{oracle} sparse patterns of four different input sequences obtained from \textit{top-k} selection over the original full attention matrix. The yellow dots indicate that the important positions in the attention matrix, while the purple region is masked out.
Figure \ref{fig:dsa_pattern} shows the sparsity patterns generated by DSA prediction. As we can see from the two figures, horizontally, the sparse attention pattern changes with different input sequences. Vertically, the predicted patterns are very close to the \textit{oracle} patterns. In our experiments, the prediction accuracy is around $85\sim95$\%.

To make sure the high performance of DSA comes from the proposed approach rather than the task itself, we further test two cases on the Text Classification dataset. Firstly, we apply a 99\% sparsity constraint on the vanilla transformer, but with a static local attention pattern. Secondly, we use a short sequence with dense attention, and let the total number of tokens in the short sequence matches with the number of important tokens in the long-sequence scenario. The results show that these two cases perform very poorly on the task, delivering a model accuracy of only 53.24\% and 54.16\% compared with 64.04\% accuracy achieved by \textbf{DSA-99\%}. This further supports our previous discussion. 

\subsection{Design Space Exploration of Prediction Path}
One of the most important design choices of DSA is the configuration of the Prediction Path. Overall, we want the predictor to accurately capture dynamic sparse patterns. However, we also want to minimize the cost of prediction while maintaining DSA model accuracy. Thus, while we involve trainable parameters for prediction, we also introduce random projection matrix $P \in \{-1, 0, 1\}^{d \times k}$ to control the prediction parameters ($\tilde{W_Q} \in \mathbb{R}^{k \times k}, \tilde{W_K} \in \mathbb{R}^{k \times k}$), and use low-precision to reduce the computation overhead. Here, we present the sensitivity results regarding different choices of the reduced dimension size and quantization precision.

\begin{table*}[t]
\centering
\vspace{-5pt}
\caption{Change of \textbf{DSA-90\%} model accuracy when sweeping random projection scale $\sigma$ and quantization precision.}
  \label{sensitivity}
	\begin{tabular}{c|c|c|c|c|c|c|c}
    \textbf{$\sigma$} &  \textbf{0.1} & \textbf{0.16}  &\textbf{ 0.2}  & \textbf{0.25} & \textbf{0.33} &\textbf{ 0.4} & \textbf{Baseline}\\
    \midrule
   \textbf{ DSA-90\%} & 65.32 & 65.25	& 65.17	& 65.46	& 65.63	& 65.54 & 65.12\\
      \midrule
      \midrule
   \textbf{Quantization}  &\textbf{Random} &  \textbf{INT2} & \textbf{INT4}  &\textbf{INT8}  & \textbf{INT16} & \textbf{FP32} & \textbf{Baseline}\\
    \midrule
   \textbf{ DSA-90\%} & 60.42 & 64.23	& 65.56	& 65.69	& 65.63	& 65.63 & 65.12 \
	\end{tabular}
	\vspace{-10pt}
\end{table*}

We first sweep over different sizes of $k$ and evaluate the accuracy of \textbf{DSA-90\%} on the LRA Text Classification task. Here, we use $\sigma=k/d \in (0,1]$ to represent the size of the predictor. A Larger $\sigma$ value indicates more prediction parameters and better representation power, but also larger computation overhead. As we can see from Table~\ref{sensitivity}, DSA demonstrates relatively stable performance with different $\sigma$ values. Even with $\sigma = 0.1$, \textbf{DSA-90\%} still achieves a slightly higher accuracy compared with vanilla transformer. We believe this is because we use predictor to indicate the positions of the important attention weights, while passing the accurate attention weights to the output. Therefore, our predictor module can tolerate highly approximate computation as long as it can capture the relative importance in the attention matrix.

\begin{figure}
  \centering
  \vspace{-10pt}
  \includegraphics[width=0.49\textwidth]{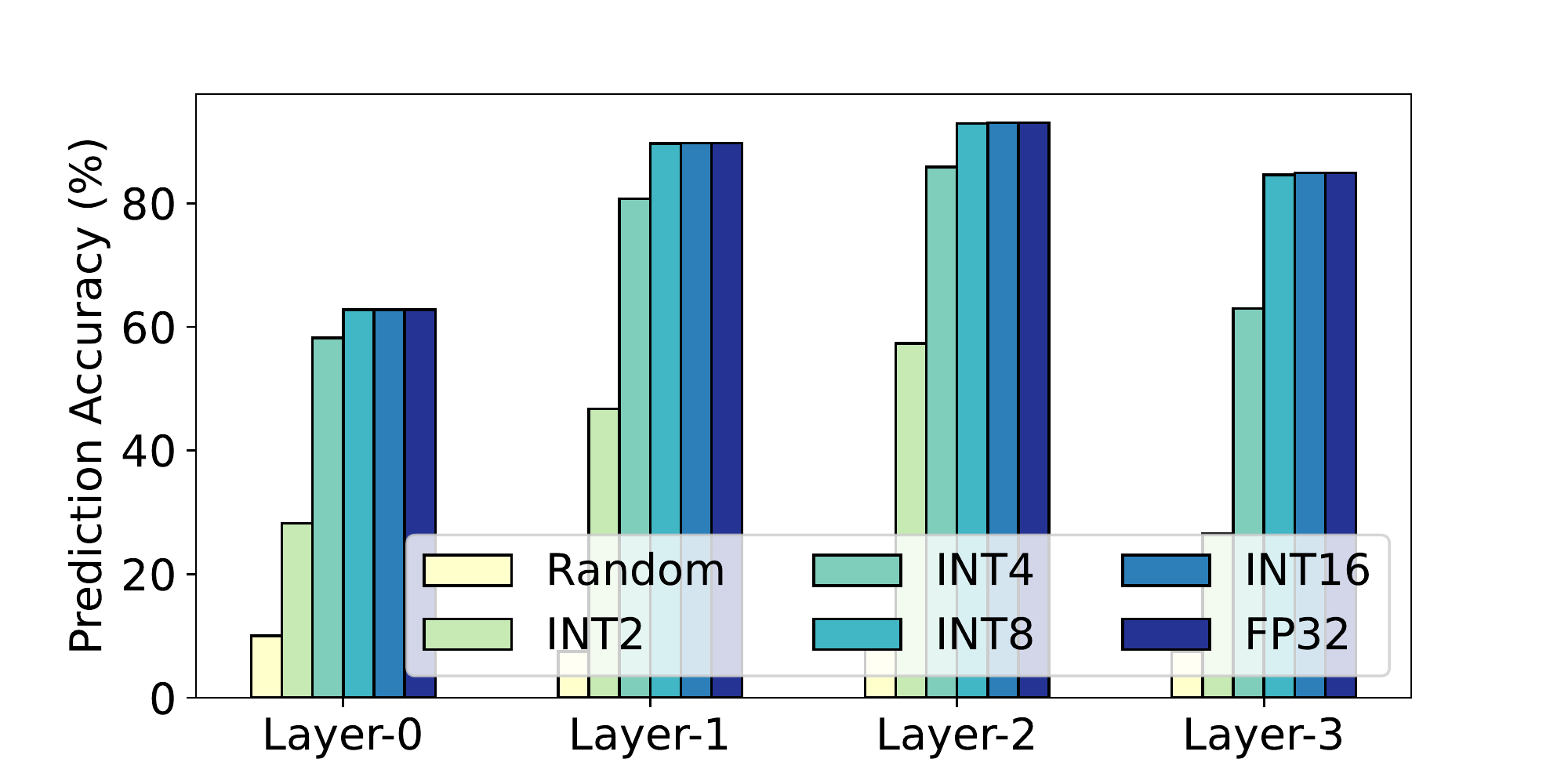}
  \vspace{-25pt}
  \caption{The prediction accuracy of DSA in a 4-layer \textbf{DSA-90\%} model with different quantization precision.}
  \label{prediction-accu}
  \vspace{-13pt}
\end{figure}

To further study the performance and the impact of the predictor, we conduct another experiment to sweep over different quantization precision, while fixing $\sigma$ to be 0.25. As shown in Table~\ref{sensitivity}, \textbf{DSA-90\%} achieves good accuracy with quantized precision as low as 4-bit. Accuracy degradation occurs when the precision further scales down to 2-bit. As we go deeper into the predictor module, we collect and show the prediction accuracy in each attention block of this 4-layer DSA model. The prediction accuracy is defined by the percentage of the correct guesses among the total number of predictions. For example, for a \textbf{DSA-90\%} model working on a sequence length of 2000, for each row of the attention matrix, the predictor will output 200 positions to be important. If 100 of these 200 locations actually matches with the $top$-$k$ results, the prediction accuracy is 50\%. As shown in Figure~\ref{prediction-accu}, the predictor is able to maintain its prediction accuracy even with 4-bit quantization. When the precision is 2-bit, the prediction accuracy suffers a significant degradation, dropping from $60\sim90$\% to $25\sim55$\%. Despite this, the overall model accuracy is acceptable, with only 0.89\% degradation compared with the baseline transformer. We believe this is because, for the binary Text Classification task, it is more crucial to capture the very few most important attentions. Although the prediction accuracy becomes lower, the most important positions are preserved and therefore maintaining the overall model accuracy. Finally, in Figure~\ref{prediction-accu} and Table~\ref{sensitivity} we include a special case of randomly selecting 10\% important positions. With this random mask applied to the model, the prediction accuracy is less than 10\%, and overall model accuracy directly drops to 60.42\%. This result supports our previous analysis.

\subsection{Model Efficiency}\label{sec:eval_efficiency}
As we mentioned earlier, DSA has the potential to significantly reduce computation and memory consumption of the self-attention layer, which is especially beneficial for deploying a long sequence transformer model at inference time. While we acknowledge that the actual runtime performance and memory footprint are largely depending on the underlying hardware implementation, in this subsection we shed light on this problem by quantitatively analyzing the cost of DSA.

\begin{figure}[!ht]
    \centering
    \vspace{-10pt}
    \includegraphics[width=0.49\textwidth]{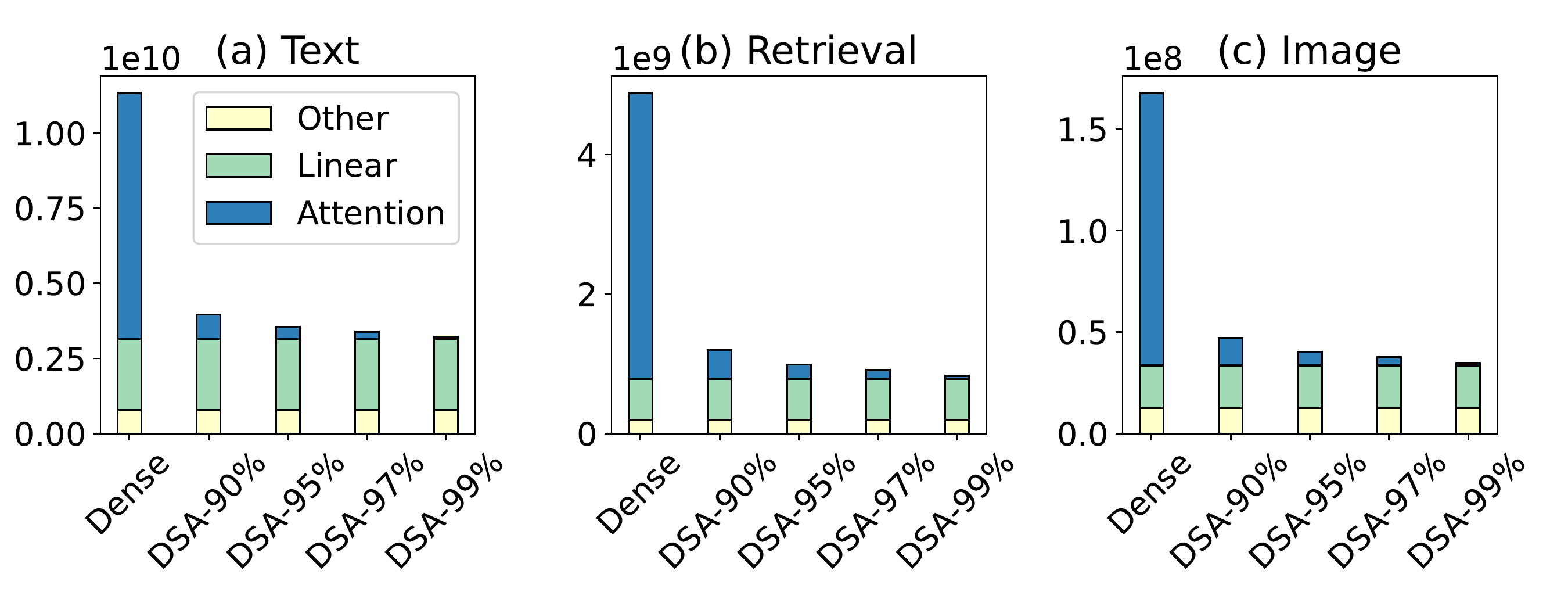} \vspace{-25pt}
    \caption{Computational cost measured in the number of MACs.}
    \label{fig:MAC}
    \vspace{-10pt}
\end{figure}

We start with presenting the number of required MAC operations for each attention layer. We use MAC number as the computational cost metric because the majority of the operations in the self-attention layer are matrix multiplications. We break down the total MAC operations into three parts: (1) Linear: General Matrix-matrix Multiplication(GEMM) for computing Query, Key, and Value. (2) Attention: GEMM for computing attention weight matrix and output Value. (3) Other: Other GEMMs inside the attention block like Feed-Forward layers. As we introduced earlier, the two GEMM operations in the part (2) scale quadratically with the sequence length, and we transform them to be SDDMM and SpMM in our DSA model to reduce both computation and memory consumption. Based on this setting, the computational cost breakdown of different models used in our LRA experiment is shown in Figure~\ref{fig:MAC}. Comparing different tasks, the tasks with longer sequence length (Text and Retrieval) are more bounded by the Attention part. The benefit of using DSA is also more significant on the 4K tasks. Comparing within each task, it is obvious that DSA model with higher sparsity ratio delivers higher computation savings. Overall, DSA achieves $2.79\sim4.35\times$ computation reduction without any accuracy degradation.

\begin{figure}[!ht]
    \centering
    \vspace{-5pt}
    \includegraphics[width=0.49\textwidth]{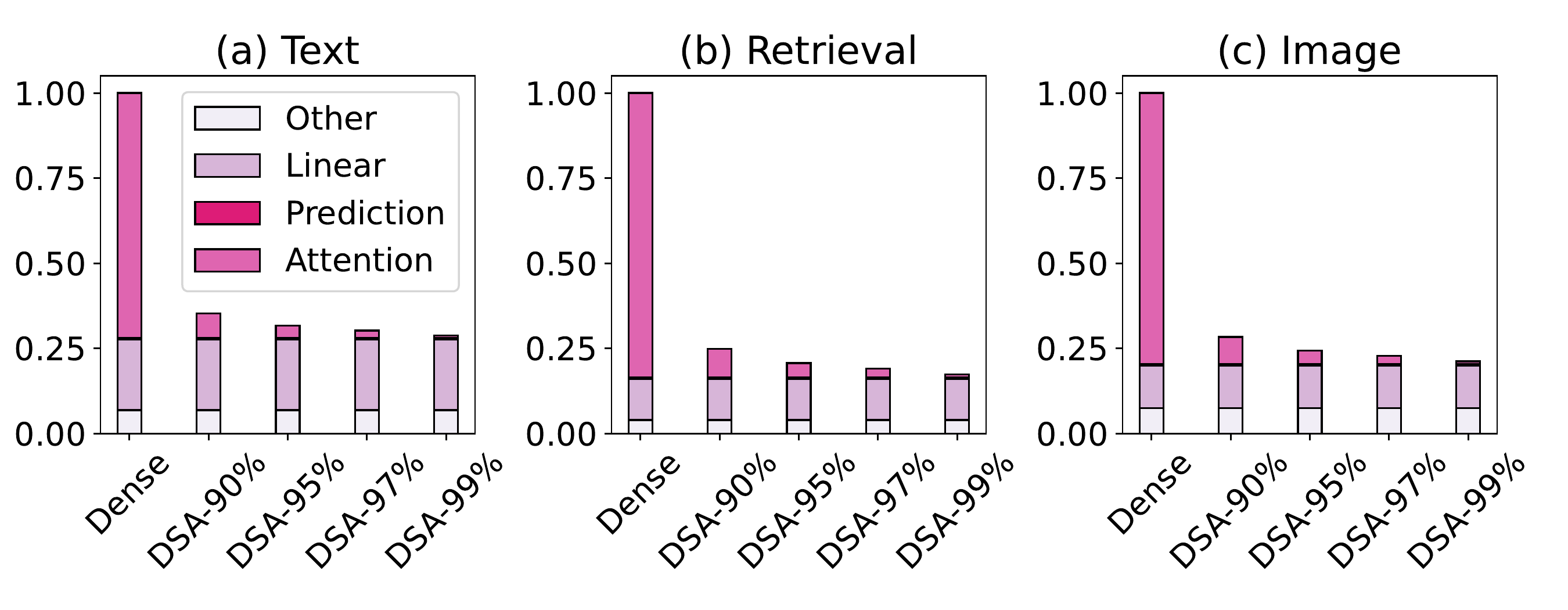} \vspace{-25pt}
    \caption{Relative energy consumption projected to vanilla transformer.}
    \label{fig:energy}
    \vspace{-20pt}
\end{figure}

Note that we do not include the computation overhead of the prediction path for generating the sparsity mask. This is because the computations conducted in prediction are in reduced precision rather than full-precision. Besides, it is inappropriate to directly project the number of low-precision MACs to the number of FP32 MACs. Therefore, we use the relative energy consumption to illustrate the overall cost of DSA-augmented attention. As shown in Figure~\ref{fig:energy}, we show the relative energy consumption of \textbf{DSA-95\%} with $\sigma=0.25$ and INT4 quantization. Each INT4 MAC's energy cost is projected to the relative factor of FP32 MAC, where the factor number is referenced from industry-level simulator~\cite{NeuroMeter} with 45nm technology. From the figure we can see that, even with the predictor overhead considered, the overall benefit is still compelling by virtue of the high dynamic sparsity ratio and low-cost prediction methodology.

\section{Algorithm-Hardware Co-design}\label{sec:hardware}

In Section~\ref{sec:eval_efficiency}, we analyze the potential of DSA in terms of reducing the total cost of Transformer model. While the estimated number of MAC operations and relative energy consumption present very promising results, it remains challenging to achieve practical speedup and energy reduction on real hardware systems. In this section, we dive deeper into this problem as we discuss the implementation of DSA on GPUs and Accelerators. Specifically, we evaluate the challenge of mapping DSA onto different platforms, and we demonstrate the flexibility of DSA to enable efficient algorithm-hardware co-designs.

\subsection{GPU Acceleration with DSA}
Given the predicted sparse patterns, we can reformulate $QK^\top$ as the sampled dense dense matrix multiplication (SDDMM) and $AV$ as the sparse matrix-matrix multiplication (SpMM). Under fine-grained sparsity, a recent work \citep{gale2020sparse} proposes SpMM and SDDMM kernel that outperforms dense GEMM kernel under $>71\%$ and $>90\%$ sparsity, respectively. Besides, \textit{cusparse}~\citep{naumov2010cusparse} also achieves practical speedup at $>80\%$ sparsity for single precision data. As we presented in Section~\ref{sec:eval}, DSA can easily deliver a sparsity ratio of more than $90\%$ with zero accuracy degradation, therefore enabling faster kernel implementations on GPUs. 

While fine-grained sparse GPU kernels are able to outperform the dense counterparts on relatively high sparsity ratios, the speedup is significantly limited due to low data reuse. Moreover, when half precision (FP16) is used for computation, above fine-grained kernels can hardly compete with GEMM kernel, as NVIDIA Tensor Core provides much higher throughput for half precision matrix multiplication. Thus, we find that the performance gain on sparse matrix multiplication can hardly mitigate the overhead of computing the prediction path in DSA, especially for half precision scenarios that commonly appeared at inference. To tackle this problem, structural dynamic sparsity can be introduced to the attention selection. Specifically, instead of selecting $top-k$ independent attention weights, we can enforce block-wise and vector-wise constraints. Also, trade-off can be made by adjusting the block size, as larger blocks deliver higher speedup but can potentially cause accuracy loss.

\begin{figure}[t]
  \centering
  \includegraphics[width=0.30\textwidth]{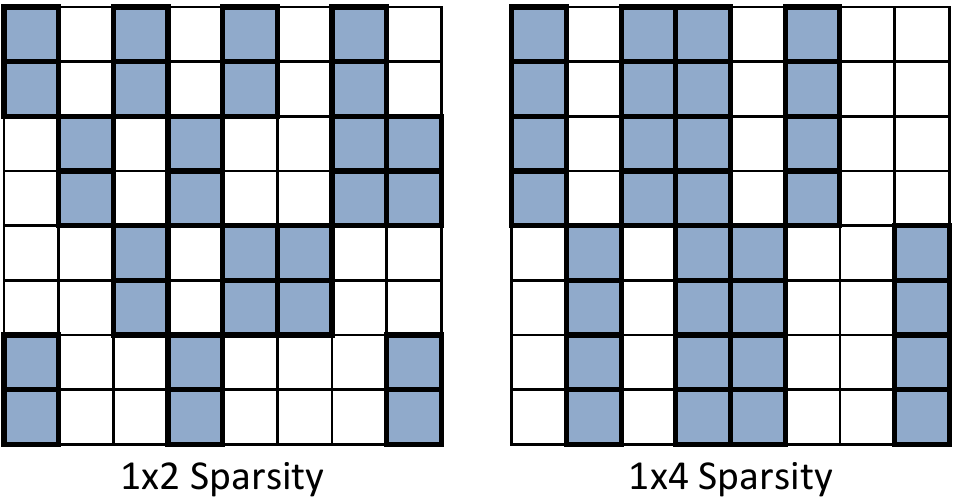}
  \vspace{-8pt}
  \caption{Column-vector sparse encoding~\cite{chen2021efficient}.}
  \label{fig:sparsity_structure}
  \vspace{-20pt}
\end{figure}

In our work, we experiment on vector sparsity using the Text Classification benchmark. As shown in Figure~\ref{fig:sparsity_structure}, we choose column-vector sparse encoding, where the attention elements are pruned in a column-vector granularity. Column-vector sparsity provides the same data reuse as block sparsity, but its smaller granularity makes it more friendly to model training~\cite{chen2021efficient}. 
Table~\ref{tab:structure_sparsity} gives the corresponding kernel speedup and model accuracy under 90\% sparsity ratio. The data type is FP16 for $1\times4/1\times8$ sparsity and FP32 for fine-grained sparsity. As we can see, DSA can be flexibly combined with different sparsity patterns, achieving practical runtime speedup on GPU while maintaining on-par model accuracy with full attention.

\begin{table}[t]
\caption{Model accuracy and kernel speedup over \textit{cuBLASHgemm}. We implement customized SDDMM/SpMM kernel for $1\times4/1\times8$ sparsity and reuse the kernel in \cite{gale2020sparse} for fine-grained sparsity. Experiments are done on NVIDIA V100 GPU.}
\vspace{5pt}
\setlength\tabcolsep{5pt}
\label{tab:structure_sparsity}
\begin{tabular}{c|c|c|c}
\hline
 Sparsity Pattern     & vec 1$\times$4 & vec 1$\times$8 & Fine-grained  \\ \hline
SpMM Speedup &   1.57$\times$   &   1.94$\times$    &   1.85$\times$ \\ \hline
SDDMM Speedup  & 0.94$\times$ &    1.15$\times$     &   1.09$\times$ \\ \hline
Accuracy(\%) & -0.02      & -0.1      & +0.5    \\ \hline
\end{tabular}
\end{table}

To shed some light on the results, we can trace back to the visualizations of the attention matrix in Figure~\ref{fig:probs}. As shown by the figure, despite the sparse and dynamic characteristics of the attention matrix, the distribution of important attention connections exhibits a certain degree of locality. For example, there exist some global tokens that attend to most of the tokens within a sequence. Therefore, some columns of the attention matrix will contain many important positions. Besides, local attention also indicates row-wise locality, as a token is likely to be influenced by its neighbors. Therefore, row-vector sparsity can be added to DSA for performance/accuracy exploration as well. While these fixed locality patterns have been well discussed in prior work~\cite{zaheer2020big, beltagy2020longformer}, DSA illustrates the dynamic distribution which motivates us to propose the prediction path to efficiently locate these important connections.

\begin{figure}[!ht]
  \centering
  \includegraphics[width=0.45\textwidth]{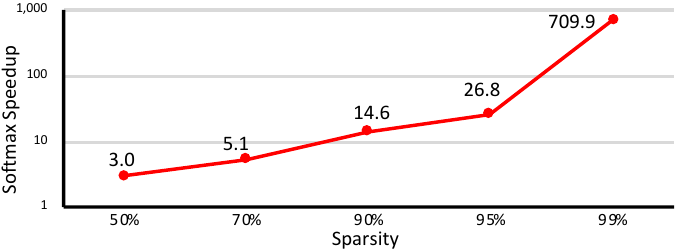}
  \vspace{-10pt}
  \caption{Speedup of softmax with different sparsity ratios.}
  \label{fig:softmax}
  \vspace{-5pt}
\end{figure}

\textbf{Sparse Softmax Computation.}
Under the long-sequence scenario, the softmax function could be a bottleneck. Let $h$, $l$, and $d$ be the number of head, sequence length, and feature dimension of each head, respectively. Our profiling result shows that with $h=8$, $l=4096$, $d=64$, softmax contributes 47\% of the total execution time of the multi-head self-attention layer. By sparsifying the attention matrix, DSA directly saves both memory access and computation consumption of the softmax function to reduce execution time. We evaluate the latency of the pytorch-implemented softmax function on NVIDIA V100 GPU. Following the configuration in Text Classification Benchmark, we set batch size=16, $h=4$, $l=2000$ and enforce different sparsity ratios. Figure~\ref{fig:softmax} shows that the reduced softmax achieves $3.0\sim709.9\times$ speedup compared with dense softmax function.

\subsection{Hardware Specialization for DSA}
While adding structural constraints can potentially benefit GPU kernel implementation, the expressive power of the model is still inevitably affected. For instance, as shown in Table~\ref{tab:structure_sparsity}, the $1\times4$ vector encoding achieves comparable accuracy with full-attention, but is lower than the accuracy of using fine-grained sparsity under the same sparsity ratio. Thus, an alternative approach is to use hardware specialization to fully exploit the potential saving from DSA.

As we know, self-attention mechanism mainly involves matrix-matrix multiplication, which can be efficiently handled with a 2D spatial array of processing elements (PEs). Prior work also proposes efficient dataflow for computing attention layer using techniques such as operator fusion, loop tiling, and loop reordering~\citep{park2020optimus, Kao2021ATTACCTQ}. With DSA, the underlying spatial array and data-flow should be adjusted accordingly. Specifically, DSA poses two architectural implications as follows.

\textbf{Multi-precision Computation.}
DSA relies on dimension reduction and quantization to control the overhead of attention prediction. Therefore, the system needs to handle both high-precision (eg., FP32/FX16) and low-precision (e.g., INT2/INT4) computations. This can be implemented either with a decoupled design or a coupled design. In decoupled architecture, standalone hardware modules are implemented for different precision \cite{liu2020duet}, e.g., using two PE arrays for low-precision and high-precision. The two modules work in a pipelined manner, where the small PE array generates sparsity information for the large PE array to skip unnecessary computations. A drawback of this type of design is that, the computation throughput is fixed for the two modules, but the relative workload between prediction and execution is task-dependent. As a result, one module may become idle time to time due to workload imbalance. On the contrary, a coupled PE array tackles this problem by using multi-precision arithmetic units~\cite{sharma2018bit}. Specifically, the computation precision of each PE is configurable, such that different sections of the PE array can be dynamically controlled to balance the relative throughput. Yet, this requires runtime configuration which makes system control to be more complicated.

\textbf{Sparsity-aware Execution.}
As mentioned above, with DSA, we can reformulate $Z=(QK^\top)V$ into a SDDMM followed by a SpMM (softmax is omitted for simplicity). In other words, the sparsity information is used as output sparsity (OS) for the first matrix multiplication and used as input sparsity (IS) during the second matrix multiplication. However, different PEs may also encounter workload imbalance due to irregularly distributed sparsity, causing low utilization. Prior work tackles this problem from multiple approaches. For example, one can enable early finished PE to switch to the execution of other elements, or apply offline sorting and online shuffling to balance the computation~\cite{song2018prediction, aklaghi2018snapea, gondimalla2019sparten}. These approaches impose different software and hardware overheads such as larger scratchpad memory, redundant memory access, higher bandwidth requirement and so on. In DSA, we use a simple and effective solution by enforcing a row-wise constraint such that different attention rows contain the same amount of important attention weights. 

\begin{figure}[!ht]
  \centering
  \vspace{-3pt}
  \includegraphics[width=0.48\textwidth]{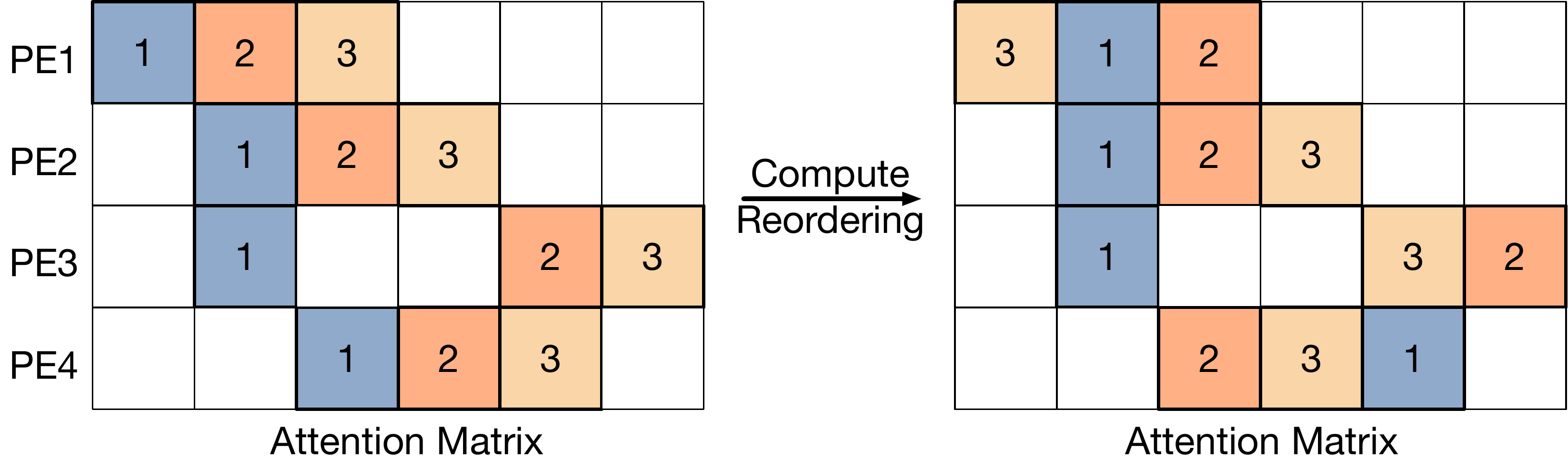}
  \vspace{-18pt}
  \caption{Using sparsity locality and compute reordering to improve data reuse.}
  \label{fig:ooo}
  \vspace{-3pt}
\end{figure}

Finally, the locality of important attention weights provides opportunities for data reuse. As shown in Figure~\ref{fig:ooo}, if multiple rows of attention matrix $A$ are computed simultaneously, the corresponding columns in matrix $K^T$ can be loaded once and shared by different PEs. Similarly, this holds true for computing matrix $Z$, as the rows in matrix $V$ can be reused. In this example, suppose four PEs work in parallel to compute the attention matrix, and each PE is responsible for one row. The colored squares are selected attentions. The numbers in the square indicate the computation order of each PE. As we can see, in the left figure, each PE computes the selected attention weights from left to right. Thus, although the sparsity distribution delivers some locality, the data reuse is bad. In contrast, if we reorder the computation within each row as shown in the right figure, then we can utilize the column locality to improve data reuse. We evaluate the benefit of computation reordering on real benchmarks. As shown in Table~\ref{tab:mem_reduction}, on the Text Classification task, the locality naturally brings $1.28\times$ memory access reduction compared with row-by-row processing, while reordering further improves this ratio to $2.54\times$. On Image Classification, the reduction ratio is $1.07\times$ without reordering and $1.37\times$ with reordering.

An important benefit of this type of \textit{out-of-order} execution is that, matrix $A$ does not need to be reshuffled and matrix $Z$ is still generated in a regular order. This granted advantage comes from the attention mechanism itself, because the whole computation process is a two-step GEMM chain. Therefore, the reordered $A$ is completely consumed during the second matrix multiplication. In contrast, exploring the same reordering in CNN would require a crossbar-like design to correctly store the output result~\cite{liu2020duet}, causing additional performance and resource overhead.

\begin{table}[h]
\vspace{-16pt}
\caption{Memory access reduction of the second matrix operand used in the multiplication of $QK^\top$ and $AV$.}
\setlength\tabcolsep{18pt}
\label{tab:mem_reduction}
\begin{tabular}{c|c|c}
\hline
 Dataflow     & Image & Text  \\ \hline
row-by-row &  $1\times$  &   $1\times$   \\ \hline
\begin{tabular}[c]{@{}c@{}}row-parallel w/o\\ compute reordering\end{tabular} &  $1.07\times$  &   $1.37\times$       \\ \hline
\begin{tabular}[c]{@{}c@{}}row-parallel w/\\ compute reordering\end{tabular}  &  $1.28\times$  &   $2.54\times$     \\ \hline
\end{tabular}
\vspace{-15pt}
\end{table}

\section{Related Work}
Transformers with the use of self-attention mechanism are difficult to scale with sequence length because of the quadratic time and memory complexity. Our paper focuses on the exploration of sparse attention patterns in Transformers. Other orthogonal approaches such as parameters sharing \citep{Gong2019Stacking} can mitigate the issue. We refer readers to a survey paper for a more comprehensive view of efficient Transformers \citep{Tay2020survey}.

\textbf{Static Sparse Patterns.} A straightforward way to exploit attention sparsity is to set static or fixed sparse patterns, such as local windows, block-wise, dilated patterns, or a combination of static patterns \citep{zaheer2020big,child2019generating,qiu2020blockwise}. However, as the sparse attention patterns are inherently dynamic depending on input sequences, those work lack the capability of capturing dynamic sparse patterns. As shown in our evaluation, the sparsity-saving trade-offs of representative methods using static sparse patterns are worse than our dynamic sparse attention approach.

\textbf{Clustering-based methods.} Building upon static block-sparse patterns, another line of research is to group similar tokens into chunks and perform local attention within chunks \citep{Kitaev2020Reformer,roy2021efficient,tay2020sparse}. The similarity function used to group tokens can be hashing, clustering, or learned sorting. However, those methods are designed for training memory reduction and impractical at inference time when operating on each sequence. The quality of grouping, e.g., convergence of clustering, is not guaranteed at long sequences, and the overhead of on-the-fly clustering is not acceptable.

\textbf{Approximation methods.} Recent work proposes to replace standard attention with forms of approximation of the attention weights \citep{wang2020linformer,katharopoulos2020transformers,choromanski2021performer,peng2021random}. While we provide a comparison in our evaluation, we regard those work out the scope of our discussion for exploring sparsity in (standard) attention. Whether using a form of approximation to replace standard attention or as we suggest to predict sparse patterns explicitly is a design choice leaving up to practitioners.

\textbf{Attention and Transformer Accelerators}
Recent work adopt algorithm and hardware co-design to reduce the cost of attention mechanism. MnnFast~\cite{jang2019mnnfast} proposes to skip the computations of $A \times V$ based on the magnitude of the calculated attention scores. This method can only benefit the second GEMM of attention layer. $A^3$~\cite{ham2020a3} introduces attention approximation to prune the unimportant attentions. However, $A^3$ involves expensive online sorting, which causes significant performance and energy overhead. ELSA~\cite{ham2021elsa} uses sign random projection to estimate the attention weights, making the approximation much more hardware efficient, but the model quality is hurt due to inaccurate approximation. In DSA, we address these limitations by simultaneously considering approximation accuracy and efficiency. Finally, SpAtten~\cite{wang2021SpAtten} proposes cascade token pruning and head pruning to reduce the cost of both self-attention block and subsequent layers. While removing several rows and columns of the attention matrix makes the operation regular and hardware-friendly, we find this constraint to be too aggressive as the locality of attention weights usually exists in small granularity.

\section{Conclusion}
In this paper, we present Dynamic Sparse Attention (DSA), a novel method that exploits dynamic sparse patterns in attention to reduce computational cost when serving Transformers. Specifically, we show that our method can achieve up to 95\% attention sparsity without model inference quality loss. Other than prior art that uses static sparse patterns in attention, our method explores dynamic sparse patterns that are inherent in attention when processing different input sequences. Instead of replacing standard attention with other variants such as low-rank approximation methods, we augment standard attention with a prediction path as the means to locate dynamic sparsity. On one hand, attention approximation can be very efficient when only used for sparsity prediction. On the other hand, the expressive power of full attention is preserved as the important attention weights from full attention are effective in model inference. Experimental results on the LRA benchmark demonstrate superior performance and model efficiency of DSA. Furthermore, we demonstrate the potential of using DSA to improve hardware performance and efficiency. With customized kernel design and structural sparsity, DSA delivers practical speedup on GPU. The algorithm benefit can be further exploited with specialized architecture, as the hardware can fully benefit from low-precision prediction, fine-grained sparse computation, and data locality.


\bibliography{main}
\bibliographystyle{mlsys2022}

\newpage
\appendix



\section{Benchmark Descriptions and Experiment Configurations} \label{app:descript}
In our experiments, we choose Text Classification, Image Classification and Document Retrieval from Long-Range Arena, while excluding Long ListOps and Pathfinder. This is because the ListOps results in LRA exhibit significant divergence without much explanation. And for Pathfinder, we are unable to reproduce the baseline results with the given training configurations from LRA.

\subsection{Text Classification}
The Text Classification task, as introduced in LRA~\citep{long-range-arena}, is a binary classification that uses real-world data to benchmark the ability of the models to deal with compositionality. IMDb review~\citep{IMDB} is selected as the dataset, which is a common choice for document classification. Moreover, to make the problem more challenging, this task takes a byte-level setup instead of the normal character-level setup for language modeling. Therefore, the model needs to learn from the unsegmented data and make compositional decisions.

For model configuration, we use the original hyperparameters given in the LRA repository \footnote{\url{https://github.com/google-research/long-range-arena}}. Specifically, the baseline transformer consists of 4 attention layers, each with 4 heads. The hidden dimension size is 256 and the positional FFN layer has a dimension size of 1024. The learning rate is 0.05 with a weight decay of 0.1. Finally, the baseline model is trained for $20K$ steps where the first $8K$ are warmup steps and the batch size is 32.

When compared with the dense baseline in Figure 2 of the full paper, the \textbf{DSA-x\%} models are obtained from fine-tuning the dense model for $5K$ steps with different levels of sparsity constraints. 
During fine-tuning, parameters from both original model and the predictor are updated simultaneously using the combination of cross-entropy loss and MSE loss. The weight factor $\lambda$ of the MSE loss is 0.01 and the learning rate is uniformly set as 0.0002.

When compared with other efficient transformers as shown in Table 1 of the full paper, we directly train the DSA prediction path from scratch. The overall training step is still $20K$, but we use the first $15K$ to train the original model and freeze the predictor module. Therefore, the first $15K$ steps are the same as training a dense baseline. After this, we jointly optimize the model and the predictor module during the last $5K$ steps with the same MSE loss factor and learning rate as above.

Finally, to limit the training cost, we set the sequence length to be 2000 for the baseline comparison and sensitivity study, while only set the length to be 4000 when comparing with other models.

\subsection{Document Retrieval}
Document Retrieval is a binary classification task that serves as a test to evaluate how well a model compresses long sequences into representations for similarity-based matching. This task uses ACL Anthology Network \citep{AAN} and aims to identify if two papers have a citation link. Similar to Text Classification, byte-level setup is used to increase the difficulty of the problem.

We use a uniform sequence length of 4000 in this task. The baseline transformer consists of 4 attention layers. Each attention layer has 4 heads, 128 hidden dimensions, and 512 FFN dimensions. The learning rate is 0.05 with a weight decay of 0.1. The model is trained for $5K$ steps with Adam optimizer and a batch size of 32. Similar to the strategy in the Text Classification task, we use fine-tuning for baseline comparison and training-from-scratch for cross model comparison. The $5K$ steps are equally divided into $2.5K$ for dense training and $2.5K$ for joint training in the training-from-scratch experiment. When jointly optimizing all the parameters, the weight factor $\lambda$ of the MSE loss is 0.01 and the learning rate is 0.0002.

\subsection{Image Classification}
The final task we include in our evaluation is image classification using CIFAR-10~\citep{CIFAR10}. Each $32\times32$ input image is flattened as a sequence of pixels. Therefore, the sequence length of this task is 1024. The input images are mapped to a single gray-scale channel where each pixel is represented with an 8-bit integer value. Following the given settings, the baseline transformer model contains one attention layer with 8 heads, 64 query/key/value hidden dimensions, and 128 FFN dimensions. 

There are in total 45,000 training samples and 15,000 validation samples. We train the model for 200 epochs with a learning rate of 0.0005 and a batch size of 128. Same as above, we use finetuning for baseline comparison and training-from-scratch for cross model comparison. The $200$ steps are divided into $150$ for dense training and $50$ for joint training in the training-from-scratch experiment. When jointly optimizing all the parameters, the weight factor $\lambda$ of the MSE loss is 0.01 and the learning rate is 0.0002.



\end{document}